\documentclass[sigconf]{aamas}  

\usepackage{booktabs}

\usepackage{pifont}
\usepackage{caption}
\usepackage{subcaption}
\usepackage{amsmath}
\usepackage{amsthm}
\usepackage{amssymb}
\usepackage{amsfonts}
\usepackage{commath}
\usepackage{nicefrac}
\usepackage{float}
\usepackage{mathtools}
\usepackage{caption}
\usepackage{subcaption}
\usepackage{enumitem}
\setlist{nosep}

\graphicspath{ {figs/} }

\newcommand{\argmax}{\mathop{\mathrm{argmax}}\limits}

\usepackage{algpseudocode,algorithm,algorithmicx}

\algrenewcommand\algorithmicrequire{\quad \;\;\textbf{Input:}}
\algrenewcommand\algorithmicensure{\quad \; \,\textbf{Output:}}

\DeclarePairedDelimiterX\Basics[1](){ #1}

\theoremstyle{plain}
\newtheorem{thm}{Theorem}
\newtheorem{lem}{Lemma}
\newtheorem{prop}{Proposition}

\theoremstyle{definition}
\newtheorem{defn}{Definition}

\newtheorem{exmp}{Example}

\theoremstyle{definition}

\setcopyright{ifaamas}  
\acmDOI{doi}  
\acmISBN{}  
\acmConference[AAMAS'18]{Proc.\@ of the 17th International Conference on Autonomous Agents and Multiagent Systems (AAMAS 2018), M.~Dastani, G.~Sukthankar, E.~Andre, S.~Koenig (eds.)}{July 2018}{Stockholm, Sweden}  
\acmYear{2018}  
\copyrightyear{2018}  
\acmPrice{}  

\begin{document}

\title{Information Design in Crowdfunding under Thresholding Policies}  


\author{Wen Shen${}^1 \;$  Jacob W. Crandall${}^2 \;$  Ke Yan${}^3 \;$  and Cristina V. Lopes${}^1$} 
\affiliation{%
  \institution{${}^1$Department of Informatics, University of California, Irvine, Irvine, CA 92617, USA\\
 ${}^2$ Computer Science Department, Brigham Young University, Provo, UT 84602, USA\\
  ${}^3$College of Information Engineering, China Jiliang University, Hangzhou 310018, China}
  }
\email{wen.shen@uci.edu, crandall@cs.byu.edu, yanke@cjlu.edu.cn, lopes@uci.edu}
\renewcommand{\shortauthors}{Shen, Crandall, Yan and Lopes}

\begin{abstract}  
Crowdfunding has emerged as a prominent way for entrepreneurs to secure funding without sophisticated intermediation. In crowdfunding,  an entrepreneur often has to decide how to disclose the campaign status in order to collect as many contributions as possible. Such decisions are difficult to make primarily due to incomplete information.  We propose information design as a tool to help the entrepreneur to improve revenue by influencing backers' beliefs. We introduce a heuristic algorithm to dynamically compute information-disclosure policies for the entrepreneur, followed by an empirical evaluation to demonstrate its competitiveness over the widely-adopted immediate-disclosure policy. Our results demonstrate that the immediate-disclosure policy is not optimal when backers follow thresholding policies despite its ease of implementation. With appropriate heuristics, an entrepreneur can benefit from dynamic information disclosure. Our work sheds light on information design in a dynamic setting where agents make decisions using thresholding policies.
\end{abstract}


\maketitle

\section*{Introduction}

Crowdfunding reinvents the way that entrepreneurs raise external funding for implementing creative ideas. It has created a rapidly growing market that contributes an annual economic impact of tens of billions of US dollars globally~\cite{yu2017crowdfunding}. Unfortunately, not all the crowdfunding campaigns are successful because most campaigns will get funded only if they have reached the fundraising goal within a deadline~\cite{short2017research}. In fact, less than $40\%$ of the crowdfunding projects reach the targeted goals and receive the funds within the campaign deadlines~\cite{short2017research}.

Mounting research has begun to investigate the determinants of the success of crowdfunding projects. Although there might be many factors (e.g., project descriptions~\cite{marelli2016makes}, product value~\cite{agrawal2014some}, geography effect~\cite{agrawal2011geography},  reward details~\cite{marelli2016makes}, entrepreneurs' reputation~\cite{kuppuswamy2015crowdfunding} and the social network effect~\cite{agrawal2014some}) that influence a campaign's success, a recent study indicates that the number of donations made by early backers of a project is often the only difference between that project being funded or not~\cite{solomon2015don}.
A substantial body of both theoretical analyses~\cite{agrawal2014some,mollick2014dynamics,alaei2016dynamic} and empirical evidence~\cite{kuppuswamy2015crowdfunding,colombo2015internal,marelli2016makes,skirnevskiy2017influence} demonstrate that the amount of early contributions has a strong positive effect on the success of crowdfunding campaigns.  These prior studies unanimously confirm the crucial role of early contributions in the success of crowdfunding projects.

There are two main reasons why the amount of early contributions matters. First, information about contributions received early in the campaign signals to potential backers the quality of the project, which in turn can trigger social learning behavior~\cite{bandura1989human} causing potential backers to also contribute to the campaign~\cite{colombo2015internal}. An empirical study on a sample of 25,058 Kickstarter projects indicates that prospective backers usually make their pledging decisions based on how much of the project goal has already been funded by others~\cite{kuppuswamy2015crowdfunding}. Second, backers who have made an early contribution are likely to circulate the information of the project to their friends or families, which may attract additional contributions~\cite{colombo2015internal,skirnevskiy2017influence}. Both rationales indicate that it is of interest to entrepreneurs to attract as many contributions from early backers as possible.
In crowdfunding, backers are often reluctant to donate in the early days of a campaign due to high uncertainty~\cite{mollick2014dynamics,alaei2016dynamic,kuppuswamy2015crowdfunding,solomon2015don,colombo2015internal}. A major source of uncertainty is the probability of success that the campaign will get funded (i.e.,  {\em Probability of Success, or PoS})~\cite{colombo2015internal,kuppuswamy2015crowdfunding}. Prospective backers are often uncertain about entrepreneurs' abilities to collect sufficient contributions to get the project funded. For instance,  $64.12\%$ of the crowdfunding projects in Kickstarter failed to reach the target goals~\cite{kickstarter2017}. A backer~\footnote{We will use ``she" to denote an entrepreneur and ``he" a backer/agent.} experiences a monetary or non-monetary opportunity cost if the fundraising goal is not achieved (and the project not funded), even if he is refunded upon the failure of the campaign~\cite{alaei2016dynamic}. 



To attract as many early contributions as possible,  an entrepreneur must take appropriate measures to coordinate backers' actions. To do this, the entrepreneur needs to have prior knowledge about the backers' arrival process, their valuation of the project (if funded), and how they estimate the probability that the campaign will be funded. However, none of this information is perfectly known to the entrepreneur. Thus, it is challenging for the entrepreneur to figure out what actions will make backers, especially early backers,  be more willing to contribute. 

If conditions permit, the entrepreneur can manipulate backers' payoffs by offering appealing discounts to early backers that face high uncertainty~\cite{ellman2015optimal,strausz2016theory}. The problem of devising allocation and payment schemes falls into the field of {\em mechanism design}~\cite{nisan1999algorithmic}.  While illuminating, it requires additional budgets and thus diminishes the entrepreneur's revenue~\cite{ellman2015optimal,strausz2017theory}. Absent from sophisticated or even unrealistic assumptions of  the backers' private types (e.g., valuation, arrival time, departure time), it is rather difficult or even unfeasible  for the entrepreneur to implement effective mechanisms~\cite{ellman2015optimal,strausz2017theory}. This is particularly the case in online settings where the entrepreneur has little knowledge about how the backers make their projections of the campaign's {\em PoS}.

Alternatively, the entrepreneur can improve backers' beliefs of the campaign's {\em PoS} by choosing what information backers see. In particular, the entrepreneur can and is permitted to voluntarily disclose the project status (i.e., how many contributions have been collected up to a given timestamp), a critical factor that influences backers' beliefs of the campaign's {\em PoS}~\cite{kuppuswamy2015crowdfunding,alaei2016dynamic}.  The problem of determining which pieces of information are disclosed to whom is called {\em information design}~\cite{taneva2015information}.


Prior work on information design has generally assumed that backers' strategic behavior was perfectly rational and that games were well-defined (e.g., signaling games)~\cite{bergemann2007information,taneva2015information,bergemann2015sequential,bergemann2016bayes,alonso2016bayesian}.  However,  studies on consumer purchasing behavior show that buyers usually follow thresholding policies to decide whether to purchase goods or not~\cite{kau1972threshold,kahneman2003perspective,zhou2005threshold}. They often buy products when the prices are no more than their reserved values. This is particularly the case when consumers face high degrees of uncertainty and have little knowledge about the environment or the future, as frequently observed in clinical decision making~\cite{pauker1980threshold}, crowdsourcing contests~\cite{easley2015behavioral},  airline ticket sales~\cite{zhou2005threshold}, online shopping~\cite{lee2005customer}, management science~\cite{su2007intertemporal}, societies of autonomous machines~\cite{shen2017regulating} and crowdfunding~\cite{mollick2014dynamics,alaei2016dynamic}. Under certain circumstances, thresholding policies are optimal policies and hence represent rational behavior~\cite{ohannessian2014dynamic}. We thus consider the scenario where backers follow thresholding policies when they decide whether to contribute to a project or not.


In this paper, we study the information design problem in which an entrepreneur voluntarily reveals the project status to backers to influence their beliefs of the project's  probability of success. Our work contributes to the state of the art in the following ways:
\begin{enumerate}
\item We show that excessive information disclosure weakly shrinks the entrepreneur's revenue. We identify conditions when immediate disclosure is optimal in crowdfunding when agents follow thresholding policies. 
We demonstrate that immediate disclosure is optimal if the funding goal has been achieved and if the project status increases monotonically by at least one contribution each time. 
\item We introduce a heuristic algorithm called {\em Dynamic Shrinkage with Heuristic Selection (DSHS)} to to help the entrepreneur make decisions on information-disclosure policies.
\item We conducted extensive simulations with real-world dataset to compare the performance of the DSHS algorithm with the widely-adopted information-disclosure policy. Experimental results demonstrate that despite its computational efficiency,  the immediate-disclosure policy is not optimal when agents follow thresholding policies. Entrepreneurs can benefit from dynamic information design with appropriate heuristics. 
\end{enumerate}

\section*{Decision Making in Crowdfunding}
\label{sec:model}

After introducing key notations, we formalize backers' decision model and the entrepreneur's optimization problem in a crowdfunding campaign.

\subsection*{Preliminaries}
\begin{figure}[h!]
\centering
\includegraphics[width=.95\linewidth]{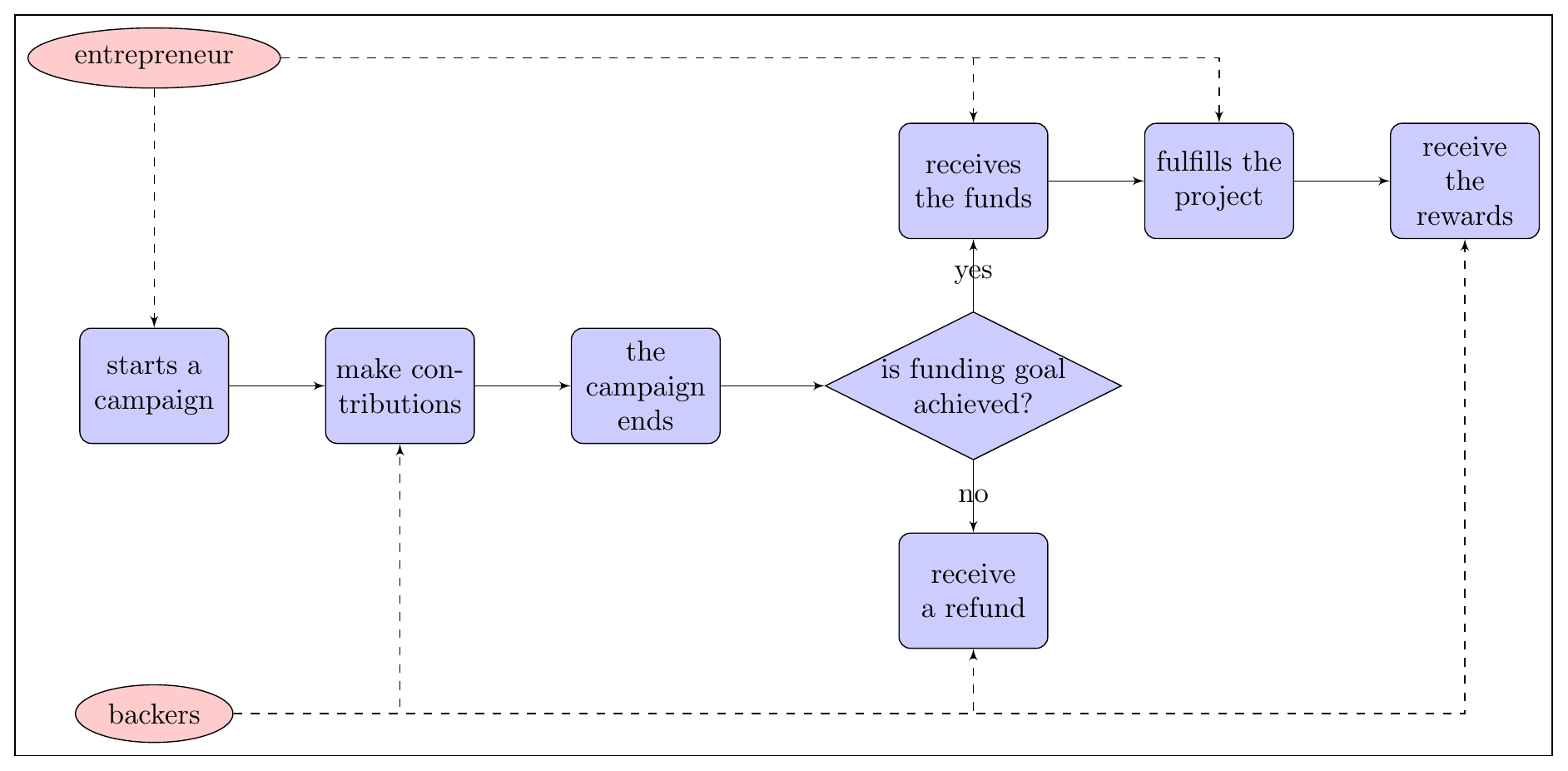}
\caption{Procedure of a typical crowdfunding campaign.}
\label{fig:timeline}
\end{figure}

We consider discrete time $t \in \mathcal{T} = \{1, 2, 3, ..., T\}$, where $T$ is the deadline for the campaign. 
Before launching the campaign, the entrepreneur must determine a fundraising goal $G$ to get funded, a deadline $T$ for reaching the goal, the number of rewards $N$, the minimal amount of contributions for a reward $P$,  and a detailed description of the project such as motivation, product, milestones, and profiles of the team. All this information is fixed and is disclosed to all the backers. See Figure~\ref{fig:timeline} for the procedure of a typical crowdfunding campaign.

After the campaign starts to accept contributions,  backers arrive at the campaign sequentially with at most one each time. This is without loss of generality because batch arrivals can be viewed as a special case where the time interval is minimal~\cite{shen2016online,alaei2016dynamic}. Let $b(t) \in \{0, 1\}$ denote the number of arrivals at time $t \in \mathcal{T}$.

At the beginning of time $t$,  the entrepreneur discloses the state of the campaign (i.e., {\em project status}) $s(k)$ to each backer $i$ that is in the campaign.   Here, $s(k) = (S_{k}, k)$ where $S_{k}$ refers to the percentage of funds that have been raised up to time $k \leq t$ ($k$ not included), with respect to the fundraising goal $G$. For simplicity, let $|s(k)| = S_{k}$.  We denote the entrepreneur's decision on information disclosure for backer $i$ at time $t$ by:
\begin{equation}
\label{eq:entredecision}
d(i,t) = (s(k), t)\; \text{ s.t. } k \in \mathcal{T},  k \leq t\;.
\end{equation}
Here, $|s(1)| = 0 $. The disclosed project status $s(k)$ must reflect the true state of the project at time $k$, which is enforced by the crowdfunding platform. In real-world crowdfunding campaigns, entrepreneurs are allowed to voluntarily disclose truthful project status~\cite{solomon2015don,alaei2016dynamic}. In our work, we assume that any information about the project status observed by the backers is directly revealed by the entrepreneur. Future work should address the scenarios  when backers have  exogenous information due to information contagion~\cite{arthur1993information}.


\subsection*{Backers' Decision Model}
It is widely known that backers' beliefs of \textit{PoS} are usually correlated with the entrepreneur's updates of the project status~\cite{kuppuswamy2015crowdfunding,marelli2016makes,alaei2016dynamic,kuppuswamy2017does}. However, the exact correlation is privately known to a backer himself only and not observed by the entrepreneur. This makes it difficult to accurately model backers' decision making process. To tackle this problem, we formalize backers' decision model using the same pattern as that in the work~\cite{ohannessian2014dynamic} by \citeauthor{ohannessian2014dynamic}, which also assumes that agents use thresholding policies.

Let $r_{i}(t, d(i,t)) \in [0, 1]$ represent high-value backer $i$'s estimate of the campaign's PoS given the report $d(i,t)$, and  $\phi_{i} \in (0,1]$ be his threshold on $r_{i}(\cdot )$ to contribute. We denote backer $i$'s decision on whether to contribute or not at time $t$ by $\alpha_{i}(t) \in \{ 0, 1\}$, where $0$ indicates \textit{Not Pledging}, and $1$ represents \textit{Pledging}.  Backer $i$'s expected utility $u_{i}$ is determined as follows:
\begin{equation}
\label{eq:backerutility}
u_{i} (t, \alpha_{i}(t), d(i, t)) =
  \begin{cases}
    c_{i} \cdot \alpha_{i}(t)  \; ,  & \quad \text{if } r_{i}(t, d(i, t)) \geq \phi_{i};\\
    0 \; , & \quad \text{otherwise.}\\
  \end{cases}
\end{equation}
Here, $c_{i} > 0$ is backer $i$'s expected utility if he contributes (i.e., $\alpha_{i}(t) = 1$) when his estimate of {\em PoS} is no less than the threshold $\phi_{i}$.  Note that $c_{i}$, $r_{i}(\cdot)$ and $\phi_{i}$ are all private information known to backer $i$ only, while his arrival and pledging behavior are observed by the entrepreneur through the platform. In practice, backer $i$ may adapt to the environment and update his threshold accordingly. In this case, his threshold $\phi_{i}$ can be treated like the upper bound of all the updated thresholds. Without loss of generality, we assume that each contributing backer pledges the same amount $P$ of fund to the project for a reward.


Backer $i$ stays at the campaign for at most $l_{i} \in \{ 1, 2, 3, .., L\}$ periods, where $l_{i}$ is known to backer $i$ only. This is without loss of generality because although backers may dynamically enter and exit the system and check the progress, these situations can be viewed as the case that the backers stay in the system for a sufficient period. 

Let $\mathcal{I}(t)$ denote the group of backers who have arrived at the campaign before or at time $t$, have at least one time period to leave and have not yet claimed a contribution. At time $t$, for each backer $i \in \mathcal{I}(t)$, his objective function is  
\begin{equation}
\label{eq:backobj}
B_{i}(t)  = \max_{\alpha_{i}(t)} u_{i}(t, \alpha_{i}(t), d(i,t)) \;  \text{ s. t. } G, T, N, P, l_{i} \; ,
\end{equation} 
where $u_{i}$  is determined by Equation~\ref{eq:backerutility}. At time $t$, backer $i$ will leave the campaign either if he claims a contribution (i.e., $\alpha_{i}(t) = 1$) or his own deadline $l_{i}$ is reached.
\subsection*{The Entrepreneur's Optimization Problem}  
In crowdfunding, the entrepreneur is interested in attracting as many contributions as possible within the deadline so that her project will get funded. Specifically, her objective is to set the disclosure policy such that the number of contributions is maximized until a given deadline $T$. 

 Let $M(t)$ denote the funds that the entrepreneur has raised up to time $t$ ($t$ included) when she uses the disclosure policy $DP(t)$. Here,  $DP(t) = (((d(i, t'))_{i \in \mathcal{I}(t')})_{t' \leq t}$. The entrepreneur's expected contributions at time $t$ is defined as follows:
\begin{equation}
\label{eq:entreobj}
M(t) = \sum_{t'=1}^{t} \sum_{i \in \mathcal{I}(t')}  \alpha_{i}(t') \cdot P \; \text{ s. t. }  G, T, N, P\; .
\end{equation}
Due to the deadline constraint, the entrepreneur's optimization problem (i.e., optimal information design) is formalized as follows:
\begin{defn}[Optimal Information  Design]
\label{def:optimalinfo}
An optimal information design in crowdfunding is to find a disclosure policy $DP_{opt}(T)$, such that $M(T)$ is maximized, i.e., $DP_{opt} = \argmax_{DP(T)} M(T)$.
\end{defn}

Due to the dynamic nature of crowdfunding, the design of disclosure policy $DP(t)$ cannot be based on backers' later decisions $(\alpha_{j}(\hat{t})_{j \in \mathcal{I}(\hat{t})})_{\hat{t} > t}$, or use later project status $(s(\hat{t}))_{\hat{t} > t}$. This constraint is called {\em No Clairvoyance}.

\section*{Optimal Information Design}
After introducing the solution concepts,  we show that excessive information weakly shrinks revenue. We further identify conditions under which immediate disclosure is optimal in crowdfunding.
\subsection*{Solution Concepts} 
In his seminal work~\cite{blackwell1953equivalent}, Blackwell formulated a partial order that is capable of comparing the quality of two pieces of information (see Theorem~\ref{thm:blackwell}). According to Blackwell's theorem, if a piece of information $\zeta_{2}$ is Blackwell-inferior to $\zeta_{1}$, then an agent will always weakly prefer $\zeta_{1}$ to $\zeta_{2}$. 
\begin{thm}[Blackwell's theorem~\cite{blackwell1953equivalent}]
\label{thm:blackwell}
Let $\zeta_{1}$ and $\zeta_{2}$ represent two pieces of information, the following conditions are equivalent:
\begin{enumerate}
\item When the agent chooses $\zeta_{1}$, her expected utility is always at least as big as the expected utility when she chooses $\zeta_{2}$, independent of the utility function and the distribution of the input.
\item $\zeta_{2}$ is a garbling of $\zeta_{1}$.
\end{enumerate}
\end{thm}

Blackwell's theorem implies two types of information: vertical information and horizontal information, which are key solution concepts used in this work.   Given two pieces of information, if one is always (weakly) preferred whatever the information receivers' types are, then they are \textit{vertical} information (see Definition~\ref{def:vertical}). If the two pieces of information are not comparable without prior knowledge about the receivers' types, then the information is \textit{horizontal} (definition omitted since it complements vertical information).
\begin{defn}[Vertical Information]
\label{def:vertical}
Given two pieces of information $\zeta_{1}$ and $\zeta_{2}$, where $\zeta_{1}\neq \zeta_{2}$,  if $\forall i \in \mathcal{I}$,  $u_{i}(\zeta_{1}) \geq u_{i}(\zeta_{2})$, then $\zeta_{1} \succsim \zeta_{2}$, where $\mathcal{I}$ denotes a  set of agents and $\succsim$,  indicating {\em preferred or indifferent to}, is independent of agent $i$'s private type.  If $\zeta_{1} \succsim \zeta_{2}$, the information is vertical.
\end{defn}


In crowdfunding, backer $i$'s estimate of the campaign's {\em PoS} (i.e., $r_{i}(t, d(i, t))$) is both time and state-dependent. Both the amount of funding (measured by $|s(k)|$) raised, and the time of the project status (denoted by $k$) are important. We identify three scenarios of vertical information. First, a higher state of project status is always more favorable if the time of the state is the same (e.g., $(20\%, 5 )\succsim (10\%, 5)$), which is obvious.  Second, the earlier report of project status is always (weakly) preferred if the project status of the two reports are the same (see Proposition~\ref{ob:earliertime}). Third, the later report of project status is always (weakly) preferred if the revenue increases by more than $P$ each time between the timestamps of the two statuses. (see Proposition~\ref{ob:grow}).  All the proofs can be found in Appendix~\ref{appendix:proofs}.
\begin{prop}
\label{ob:earliertime}
An earlier report of project status is always weakly preferred  if the project status of the two reports are the same. Formally,  given project status $s(k_{1})$ and $ s(k_{2})$, where $k_{1}< k_{2}$, $|s(k_{1})| = |s(k_{2})|$,   we have $\forall t \geq k_{2}, \forall i \in \mathcal{I}(t)$:  $(s(k_{1}), t) \succsim (s(k_{2}), t)$.
\end{prop}
\begin{prop}
\label{ob:grow}
A later report of project status is always weakly preferred if the revenue increases by more than $P$ each time between the period of the two statuses. Given status reports $\varepsilon_{1}:  (s(k_{1}), t_{1})$ and $\varepsilon_{2}: (s(k_{2}), t_{2})$, where $k_{1} < k_{2},  t_{1} < t_{2}$ and $ |s(k_{2})| - |s(k_{1})| \geq (k_{2}-k_{1}) \cdot P/G$, we have $\forall t \geq k_{2}, \forall i \in \mathcal{I}(t)$: $\varepsilon_{2} \succsim \varepsilon_{1} $.
\end{prop}

If the conditions of vertical information cannot be identified, the information is horizontal (see Example~\ref{ex:horizontal}). Without prior information about backers' private types (e.g., arrival process, valuation, the estimate of the campaign's {\em PoS}, and the correlation between them), it is not feasible for the entrepreneur to identify optimal information design.  However, an effective disclosure policy should capture both the vertical  and  horizontal component, making the information design problem particularly challenging for the entrepreneur.
\begin{exmp}
\label{ex:horizontal}
Given $T= 30, P = 0.1G, |s(10)| = 30\%$ and $|s(15)| = 40\% $,  without prior knowledge about backer $i$'s projection of {\em PoS} (i.e., $r_{i}(\cdot)$), it is unclear which project status is more favorable by $i$. This is because $|s(15)| - |s(10)| = 0.1< 0.1 \cdot (15-10)$.
\end{exmp}

\subsection*{Excessive Disclosure Shrinks Revenue}
Given two project status reports,  if their partial order can be identified according to Proposition~\ref{ob:earliertime} and~\ref{ob:grow}, then the entrepreneur only needs to disclose the one with higher order. This is because revealing the low-order
report does not increase the chance that backers contribute to the campaign (see Lemma~\ref{lem:addivertical}). 
\begin{lem}
\label{lem:addivertical}
If the order of two project status reports can be identified, the low-order report does not increase the change of backers' contribution. Formally, given project status reports $\varepsilon_{1}:  (s(k_{1}), t_{1})$ and $\varepsilon_{2}: (s(k_{2}), t_{2})$, if $\forall t \geq \max\{k_{1},k_{2}\}, k_{1} \neq k_{2}, \forall i \in \mathcal{I}(t): \varepsilon_{2}\succsim \varepsilon_{1}$,   we have : 
\begin{equation}
E(\alpha_{i} =1 | \varepsilon_{2}) \geq E(\alpha_{i} =1 | (\varepsilon_{1}, \varepsilon_{2})) \; ,
\end{equation}
where $E(\alpha_{i}=1 | \varepsilon)$ denotes the expectation that backer $i$ contributes to the campaign given the information $\varepsilon$.
\end{lem}
If the partial order of the two reports cannot be identified, the entrepreneur should also refrain from disclosing additional information. Depending on how backers estimate the campaign's {\em PoS}, revealing more information than necessary can decrease the revenue. The reason is that excessive information disclosure can decrease backers' projections of the {\em PoS} (See Lemma~\ref{lem:addihorizontal}). 
\begin{lem}
\label{lem:addihorizontal}
If the partial order of the two project status reports cannot be identified, excessive information disclosure weakly decrease backers' projections of \textit{PoS}. Formally, given project status reports $\varepsilon_{1}:  (s(k_{1}), t_{1})$ and $\varepsilon_{2}: (s(k_{2}), t_{2})$ where $k_{1} \neq k_{2}$,  if the partial order of the two cannot be identified by the entrepreneur, then $\forall t \geq \max\{k_{1}, k_{2}\}, \forall i \in \mathcal{I}(t)$, we have: 
\begin{equation}
r_{i}(t, (\varepsilon_{1}, \varepsilon_{2})) \leq  \max \{r_{i}(t, \varepsilon_{1}),  r_{i}(t, \varepsilon_{2})\} \; .
\end{equation}
\end{lem}
In our model of crowdfunding, excessive information disclosure weakly diminishes a backer's willingness to contribute (see Theorem~\ref{thm:excessive}) and thus weakly shrinks the revenue as well as the entrepreneur's ability to implement optimal information-disclosure policies. To collect as many contributions as possible, the entrepreneur should not disclose more information about the project status than necessary. 
\begin{thm}
\label{thm:excessive}
Excessive information disclosure weakly diminishes the chance that a backer will contribute. Formally, given two project status reports $\varepsilon_{1}:$  $(s(k_{1}), t_{1})$ and $\varepsilon_{2}: (s(k_{2}), t_{2})$ where $k_{1} \neq k_{2}$,  let $\alpha_{i}^{'}$ denote backer $i$'s decision on pledging if given report either $\varepsilon_{1}$ or $\varepsilon_{2}$, and $\alpha_{i}^{''}$ denote his decision on contribution if given $(\varepsilon_{1}, \varepsilon_{2})$.  $\forall t \geq \max\{k_{1}, k_{2}\}, \forall i \in \mathcal{I}(t)$, we have: $E(\alpha_{i}^{'}=1)  \geq E(\alpha_{i}^{''}=1)$.
\end{thm}

\subsection*{Immediate Disclosure is not Always Optimal}
The immediate-disclosure policy (see Definition~\ref{def:immediate}) is widely adopted by entrepreneurs on major crowdfunding platforms (e.g., \textit{Kickstarter}, \textit{Indiegogo}) due to its ease of implementation~\cite{alaei2016dynamic}.  It is thus important to investigate if immediate disclosure is optimal.

 \begin{defn}[Immediate Disclosure]
\label{def:immediate}
An immediate-disclosure policy always reveals the current project status to all the backers in the campaign. Formally let $DP_{im}$ denote immediate disclosure, we have $DP_{im}(t) = (((d(j, t'))_{j \in \mathcal{I}(t')})_{t' \leq t} \text{ s.t. }  d(j, t')=(s(t'), t')$.
\end{defn}

If the entrepreneur and the backers have identical information, immediate disclosure is optimal~\cite{rayo2010optimal,kamenica2011bayesian,au2015dynamic}. It still holds if the entrepreneur has some unique information provided that such information does not affect the backers' decisions. Unfortunately, in our model, all information about the campaign (e.g., $G, T, P, N$) except the project status is known to both the entrepreneur and the backers. Backer $i$'s estimate of {\em PoS} (i.e., $r_{i}(t, d(i, t))$) is influenced by the project status $s(k)$ that the entrepreneur reveals. Thus, it is critical for the entrepreneur to identify conditions when immediate disclosure is optimal.

From Proposition~\ref{ob:grow}, we see that to improve backer $i$'s belief of the campaign's {\em PoS}, the entrepreneur should always disclose the project status that is preferred by all the backers if available. Otherwise, the information design is not optimal. With this intuition in mind, we show that before the campaign reaches the fundraising goal, immediate disclosure is optimal if and only if the project status increases monotonically in time by at least one contribution each time (see Lemma~\ref{prop:immediate}). This condition characterizes the relationship between the growth rates of the revenue and the maximum possible arrival rate of the backers. Though the entrepreneur  does not have prior knowledge of the backers' types (e.g., beliefs, thresholds), she can observe the progress of the project and determine if immediate disclosure is optimal given the tracking record of project status. 
\begin{lem}
\label{prop:immediate}
Before the campaign reaches the fundraising
goal, immediate disclosure is optimal if and only if the
project state increases monotonically in time by at least one contribution
each time.  Formally, if $M(t) < G$, then we have: 
\begin{equation*}
\label{eq:dpim}
DP_{im}(t) = \argmax_{DP(t)} M(t) \iff \forall t' \leq t:  |s(t')| - |s(t' -1)| \geq  P/G.
\end{equation*}
\end{lem}
After successfully reaching the funding objective, it is certain that the campaign will get funded, so immediate disclosure is optimal (see Lemma~\ref{prop:afterimmediate}).
\begin{lem}
\label{prop:afterimmediate}
 After the campaign reaches the fundraising
goal, Immediate disclosure is optimal. Formally, if $M(t) \geq G$, then we have: $DP_{im} = \argmax_{DP(t)} M(t)$.
\end{lem}

Lemma~\ref{prop:immediate} shows that immediate disclosure is  not always optimal during the crowdfunding campaign when backers follow a thresholding policy. According to Definition~\ref{def:optimalinfo} and Equation~\ref{eq:entreobj}, in order to compute the optimal solution, the entrepreneur must have a prior knowledge of the sequence of decisions $((\alpha_{i}(t))_{i \in \mathcal{I}(t)})_{t \in \mathcal{T}}$ in advance.  However, such assumption violates the  \textit{No Clairvoyance} constraint and is not implementable in practice.

\section*{Dynamic Information Design}
\label{sec:dynamicinfodesign}

Instead of restricting our attention to optimal information design, we introduce a heuristic algorithm, called {\em Dynamic Shrinkage with Heuristic Selection (DSHS)}, to help the entrepreneur make decisions on information disclosure.

{\em DSHS} treats the two conditions separately: before and after project success (see Algorithm~\ref{alg:DSHS}) . Before the campaign reaches the fundraising goal, the algorithm determines the disclosure policy according to two processes: \emph{dynamic shrinkage} (see Algorithm~\ref{alg:exclude}) and \emph{heuristic selection}. After the campaign reaches the fundraising goal (if it happens), the algorithm discloses information immediately.
\begin{algorithm}
  \caption{{\em DSHS}
    \label{alg:DSHS}}
  \begin{algorithmic}[1]
   \Require{ $t$ - time; $s(t)$ - project status at time $t$;  $\mathcal{I}(t)$- backers in the campaign.}
   \Ensure{$ (d(i, t))_{i \in \mathcal{I}(t)} $- the entrepreneur's decisions on information disclosure for backers in the campaign at time $t$.}
\If{$t \leq T$}
\For{each backer $i \in \mathcal{I}(t)$}
\If{ current revenue $M(t) < G$} \Comment{before success}
\State Include all the available project status  into $H_{i}(t)$
\State Sort $H_{i}(t)$ in the ascending order of $|s(k)|$ 
\State Remove the  least promising candidates in $H_{i}(t)$ 
\State Select the project status $s(k_{sel})$ using heuristics
\State Finalize disclosure decision $d(i, t) \gets (s(k_{sel}), t)$
\Else \Comment{after success}
\State Disclosure current status, i.e.,  $d(i,t) \gets  (s(t), t)$

\EndIf
\State  Update revenue $M(t+1) = M(t) +\alpha_{i}(t) \cdot P$ 
\State Update project status $ s(t+1) = M(t+1)/G$
\EndFor
\EndIf
  \end{algorithmic}
\end{algorithm}
\subsubsection*{Dynamic Shrinkage}
In the dynamic-shrinkage process,  {\em DSHS} ranks all the available choices,  and removes the least promising choices which are less preferred by the backers according to Propositions~\ref{ob:earliertime} and \ref{ob:grow}. By doing so, the entrepreneur avoids excessive information disclosure that weakly shrinks revenue. 

Initially,  {\em DSHS} includes all the project status disclosures $s(k)$ since the last disclosure for backer $i$ into a set $H_{i}(t)$. That is, 
\begin{equation}
H_{i} (t) \gets \{s(k)\}_{k \in \{k_{0}, k_{0}+1,..., t\}}  \text{ s.t. }  d(i, t') = (s(k_{0}), t') \;,
\end{equation}
where $t' \in \{1, 2, ..., t\}$. It then sorts $H_{i}(t)$ in the ascending order of $|s(k)|$. This sorting problem can be easily solved by \emph{Quicksort}~\cite{hoare1962quicksort} with time complexity $O(|H_{i}(t)| \log |H_{i}(t)|)$. Since $|H_{i}(t)| \leq T$, the worst-case complexity for the function is $O(T \log T)$.

After the sorting process,  {\em DSHS} removes the least promising candidates through the function {\em SHRINK} (see Algorithm~\ref{alg:exclude}).  While there are at least two choices available, the {\em SHRINK} algorithm removes the project status with later time if the two statuses have the same progress (see line 4, Algorithm~\ref{alg:exclude}) according to Proposition~\ref{ob:earliertime}. This process is equivalent to removing duplicates in a sorted array, which can be solved in $O(T)$ time. Given two project statuses, if they satisfy the relation in Proposition~\ref{ob:grow}, then the algorithm removes the project status with the earlier time (see line 7, Algorithm~\ref{alg:exclude}) . This step takes $O(T\log T)$ time in the worst case. The algorithm does nothing if only one disclosure strategy exists.
\begin{algorithm}
  \caption{{\em SHRINK}
    \label{alg:exclude}}
  \begin{algorithmic}[1]
   \Require{$H$- sorted project status disclosures}
   \Ensure{$H' $-remaining status disclosures after shrinkage}
\If{$|H| \geq 2$}
\While{$s(k_{1}), s(k_{2}) \in H, k_{1} < k_{2}$}
\If{$|s(k_{1})| = |s(k_{2})|$} 
\State $H \gets H \setminus \{s(k_{2})\}$\Comment{By Proposition~\ref{ob:earliertime}}
\EndIf
\If{$|s(k_{2})| - |s(k_{1})| \geq (k_{2} - k_{1}) \cdot P/G$}
\State $H \gets H \setminus \{s(k_{1})\}$\Comment{By Proposition~\ref{ob:grow}}
\EndIf
\EndWhile
\EndIf
\State $H' \gets H$
  \end{algorithmic}
\end{algorithm} 

\subsubsection*{Heuristic Selection}
After the shrinkage process, if there are still at least two choices available (i.e., $|H_{i}(t)| \geq 2$), then the remaining set $H_{i}(t)$ is horizontal. The entrepreneur has to select some $s(k_{sel}) \in H_{i}(t)$ to attract as many contributions as possible. This optimization problem is similar with the renowned \textit{restless bandit problem}~\cite{whittle1988restless}, which is not solvable due to incomplete information. However, simple heuristics such as {\em random selection}, {\em greedy selection},  $\epsilon${\em -greedy exploration},  and {\em softmax exploration} can be used to produce acceptable results. See Appendix~\ref{apendix:algo} for details of each algorithm. 

We further introduce a meta algorithm (See Algorithm~\ref{alg:meta}). The intuition is that the algorithm can improve the quality of decisions by only using the experts that have a satisficing performance for producing the final results~\cite{crandall2014towards}. Besides, we take an ensemble approach to calculate the final selection instead of directly applying the results produced by the selected experts.  The benefit is that the algorithm can further reduce potential performance loss due to biases of a single individual expert~\cite{shen2013ensemble,kuncheva2003measures}.

Before describing the meta algorithm, we first introduce the notations used. Let $X$ denote the set of experts, where $x \in X$ is one of the four heuristics.  We write $z^{t}(x)$ for expert $x$'s expected revenue at time $t$ and write $w^{t}(x) $ for expert $x$'s revenue prospect.  Here, $z^{t}(x)$ is computed by: $z^{t}(x) = \sum_{s(k) \in H'} \sum_{i \in \mathcal{I}(t)} \rho ^{t}_{x}(s(k)) \cdot \Upsilon_{x}(s(k), i, t)$,  where $\rho ^{t}_{x}(s(k))$ denotes the probability that $s(k)$ is selected as the targeted project status by expert $x$ at $t$, and $\Upsilon_{x}(s(k), i, t)$ is the entrepreneur's expected increase of revenue given $(s(k), t)$ for backer $i \in \mathcal{I}(t)$ by using expert algorithm $x$.  Details of computing $\Upsilon_{x}(s(k), i, t)$ for each expert $x$ is described in Appendix~\ref{apendix:algo}. Initially, $w^{t} = \max_{x \in X} \{z^{t}(x)\}$. 
 
Each time the algorithm selects a subset $X'$ of experts whose expected revenue is higher than a learned threshold---the minimum learned prospect $w^{t}(x)$. This step eliminates the experts that fail to produce better expected revenue than the threshold. The algorithm then performs a majority vote from the results generated by each expert $x \in X'$. The selected project status $s(k_{sel})$ is the one with the most votes. Ties are broken by choosing the result generated by the expert with the highest $z^{t}(x)$. This step aims to improve the robustness of selection by reducing the performance loss caused by biases of a single expert.

When a new expert algorithm is selected, the prospect for the expert is updated by $w^{t}(x) = (1-\sigma) q^{t}(x)  + \sigma w^{t-\delta}(x)$. Here, $\delta$ is the number of periods that expert $x$ has been used,  and $\sigma \in [0, 1]$ is the learning rate ( $\sigma = 0.9$ in our paper).  $q^{t}(x)$ is the entrepreneur's average revenue gain per time by using expert algorithm $x$ in the last $\delta$ periods. It is calculated by: $q^{t}(x) =  \sum_{t' = t-\delta}^{ t} \sum_{j \in \mathcal{I}(t')} \frac{\alpha_{j}(t')}{\delta}$.
\begin{algorithm}[ht!]
  \caption{{\em META}
    \label{alg:meta}}
  \begin{algorithmic}[1]
   \Require{$H'$-remaining status disclosures after shrinkage;  $X$-the set of experts}
   \Ensure{$s(k_{sel}) $-the selected project status disclosure}
\State Compute $z^{t}(x) \text{ for } x \in X$
\State Initialize $w^{t}(x) = \max_{x \in X} z^{t}(x)$
\While {$t < T$}
\State $X' = \{x: z^{t}(x) \geq \min w^{t}(x)\}$
\State Perform a majority vote for $s(k) \in H_{X'}$
\State Select $s(k_{sel})$ as the $s(k)$ with the majority rule
\State Update $w^{t}(x) = (1-\sigma) q^{t}(x)  + \sigma w^{t-\delta}(x)$ if a new expert $x$ is selected
\State Update $q^{t}(x)$ and  $z^{t}(x)$ for each $x \in X$
\EndWhile
  \end{algorithmic}
\end{algorithm} 

{\em DSHS}  is highly flexible in the sense that it allows the entrepreneur to easily customize both the shrinkage process and the selection process with different methods.
\section*{Empirical Evaluation}
This section describes the experimental settings and the results.
\subsubsection*{Experimental Setup} 
We collected the campaign data from a randomly selected subset of Kickstarter\footnote{https://www.kickstarter.com} projects using a web crawler. The dataset contains 1,569 projects which satisfy the following conditions: (1) they were all-or-nothing, reward-based campaigns; (2) the campaigns lasted for exactly 1,440 hours (60 days) with both the starting time and the ending time falling between 07/15/2016 and 10/15/2016.  Each campaign includes hourly project status, the fundraising goal, the deadline, the minimal amount of contribution, and the number of contributions every hour. The data samples allow us to mimic the operation of real-world crowdfunding projects when the underlying factors and correlations that impact them are yet to be identified~\cite{mollick2014dynamics,alaei2016dynamic}.


%
Most of the projects on Kickstarter offer several tiers of perks for the backers to choose. We only selected the early bird pledges and the regular pledges that would offer a product to the backers to compensate backers' financial support. We adjusted the funding goal for each project accordingly. The early bird pledges were proportioned to the regular pledges. For instance, an early bird pledge with value $8$ is equivalent to $0.8$ regular pledge with value $10$.
 
Due to Kickstarter's API constraints, we were unable to track the number of backers who visited the campaign per hour. We simulated backers' arrivals using \emph{Poisson}~\cite{ross1996stochastic} distribution for each project.  The arrival process was independently and identically distributed with a mean $\vartheta(t)$ across time, where $t \in \mathcal{T}$.  The mean of $(\vartheta(t))_{t\in \mathcal{T}}$ was 0.1 (consistent with the empirical arrival rates of backers in crowdfunding projects~\cite{marwell2015competing}).  

We used the anticipating random walk~\cite{alaei2016dynamic} model for simulating backers' projections of {\em PoS} because it is tailored for computing backers' estimates of {\em PoS} in crowdfunding. For each project, the backers' valuation of a reward was \emph{Gaussian}~\cite{ross1996stochastic}  distributed with a mean equivalent to the value of the reward $P$ and a randomly selected standard deviation ranging 0.05 of the mean to 0.5 of the mean. 


We performed six groups of experiments:  immediate disclosure ({\em immediate}),  and {\em DSHS} with five heuristics ({\em random}, {\em greedy}, {\em $\epsilon$-greedy},   {\em softmax}, and {\em meta}) .  Each group was run 30 times with the same 2.9GHz quad-core machine.

\subsubsection*{Results} 
\begin{figure}[h!]
\centering
\begin{subfigure}[b]{.48\linewidth}
  \includegraphics[width=\linewidth]{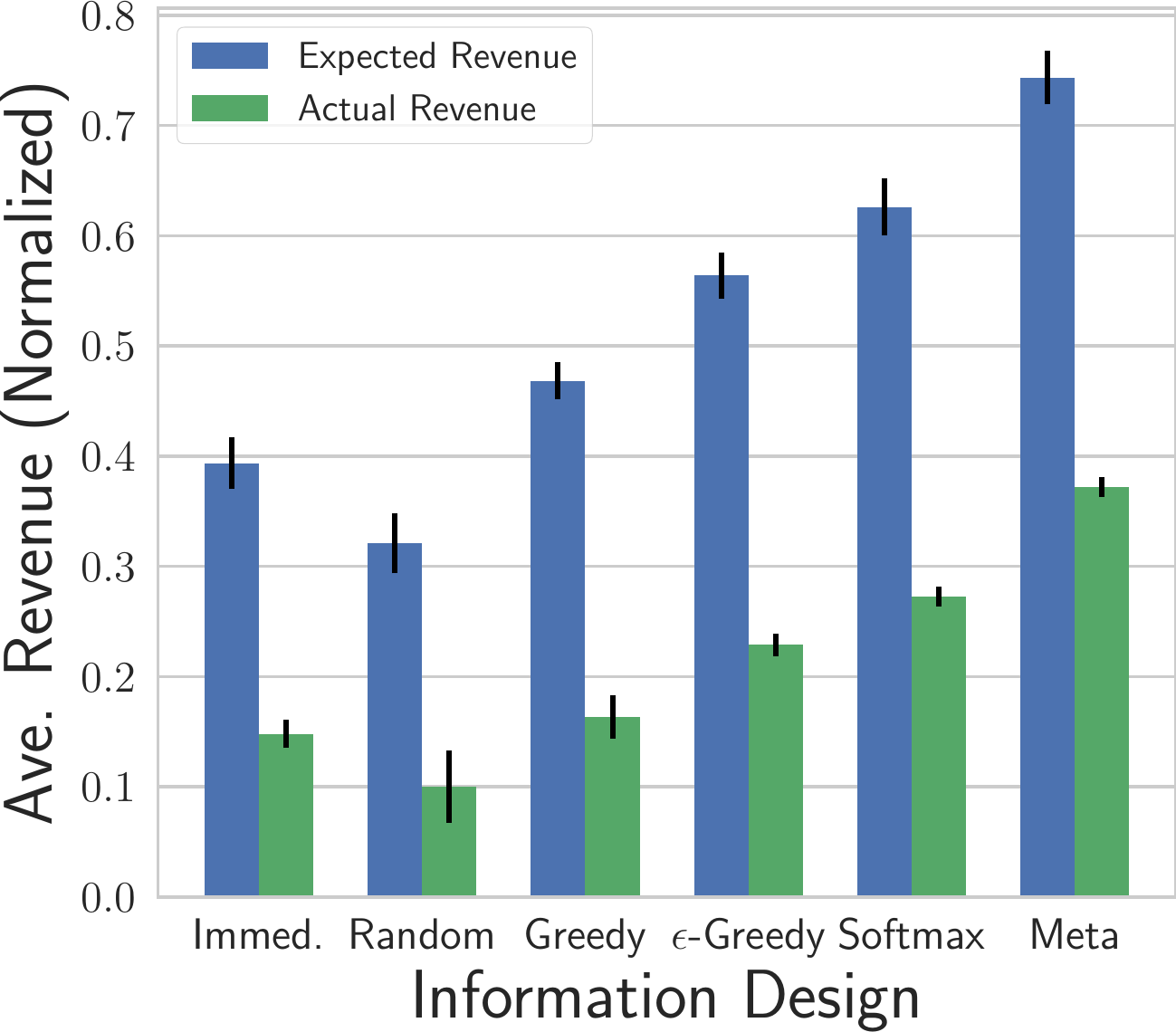}
  \caption{Overall revenue ($t=1440$)}
  \label{fig:revenueoverall}
\end{subfigure}
\begin{subfigure}[b]{.48\linewidth}
  \includegraphics[width=\linewidth]{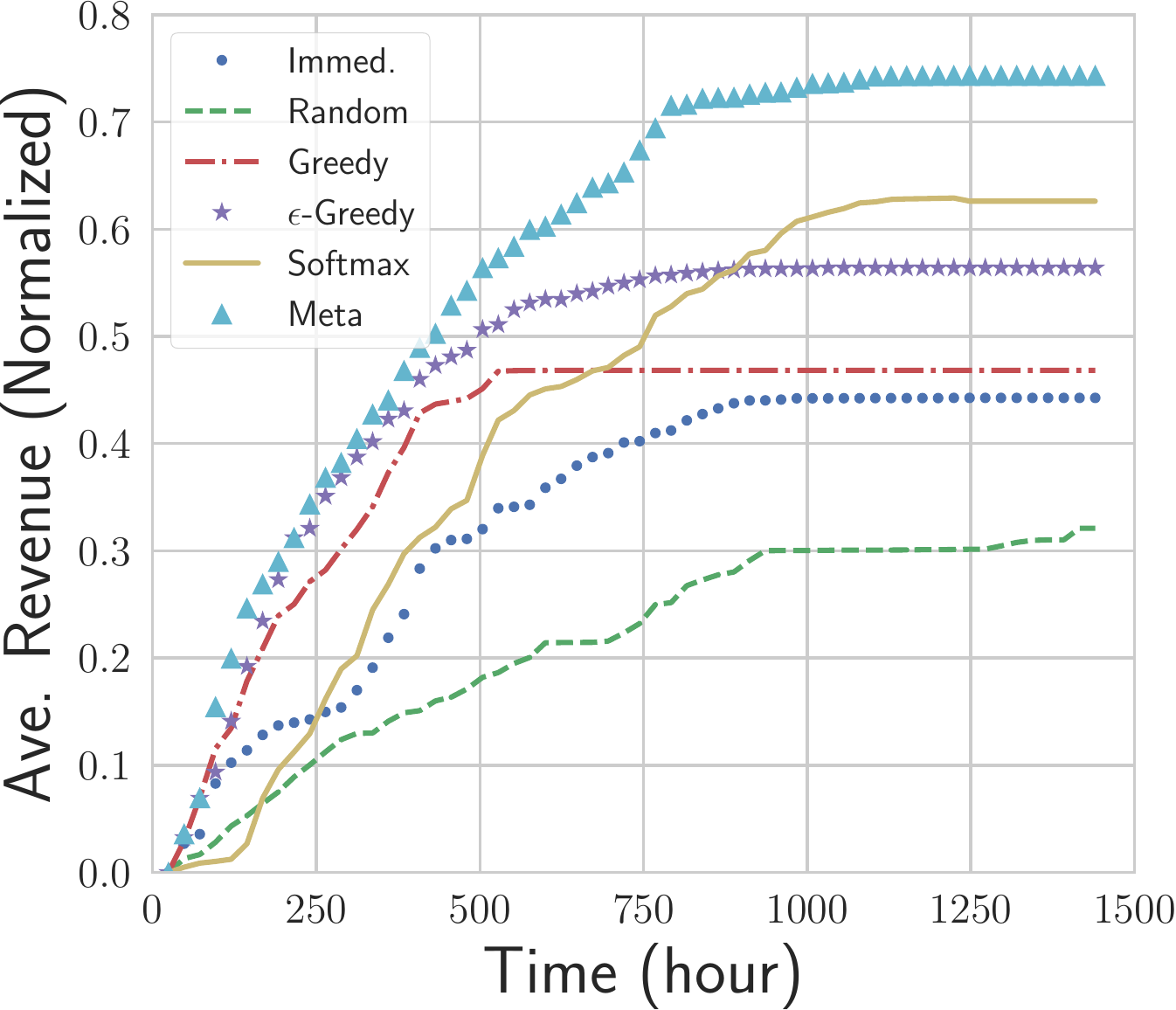}
    \caption{Expected revenue over time}
      \label{fig:revenueovertime}
\end{subfigure}
\begin{subfigure}[b]{.48\linewidth}
  \includegraphics[width=\linewidth]{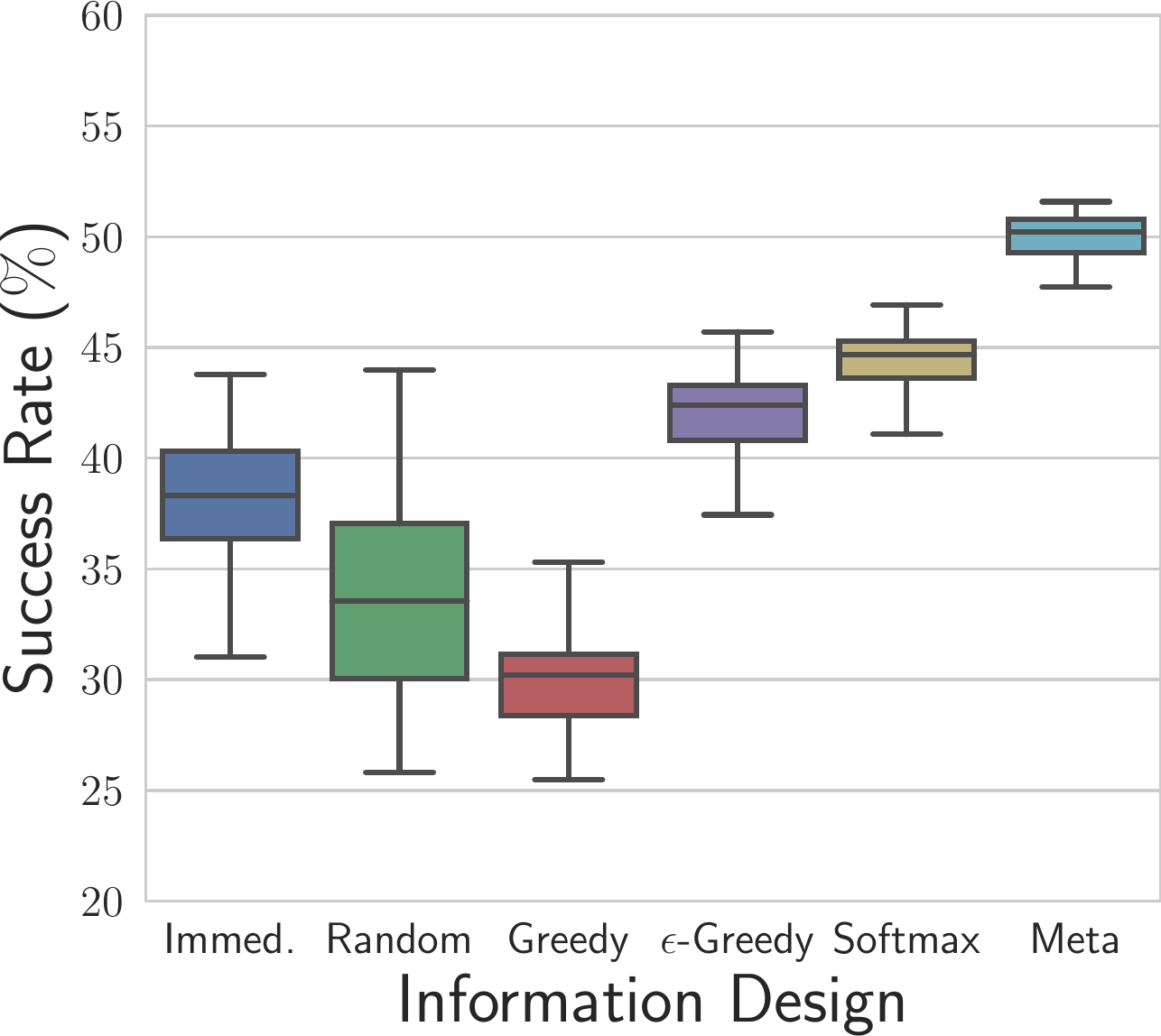}
    \caption{Project success rate ($\%$)}
     \label{fig:successrate}
\end{subfigure}
\begin{subfigure}[b]{.48\linewidth}
  \includegraphics[width=\linewidth]{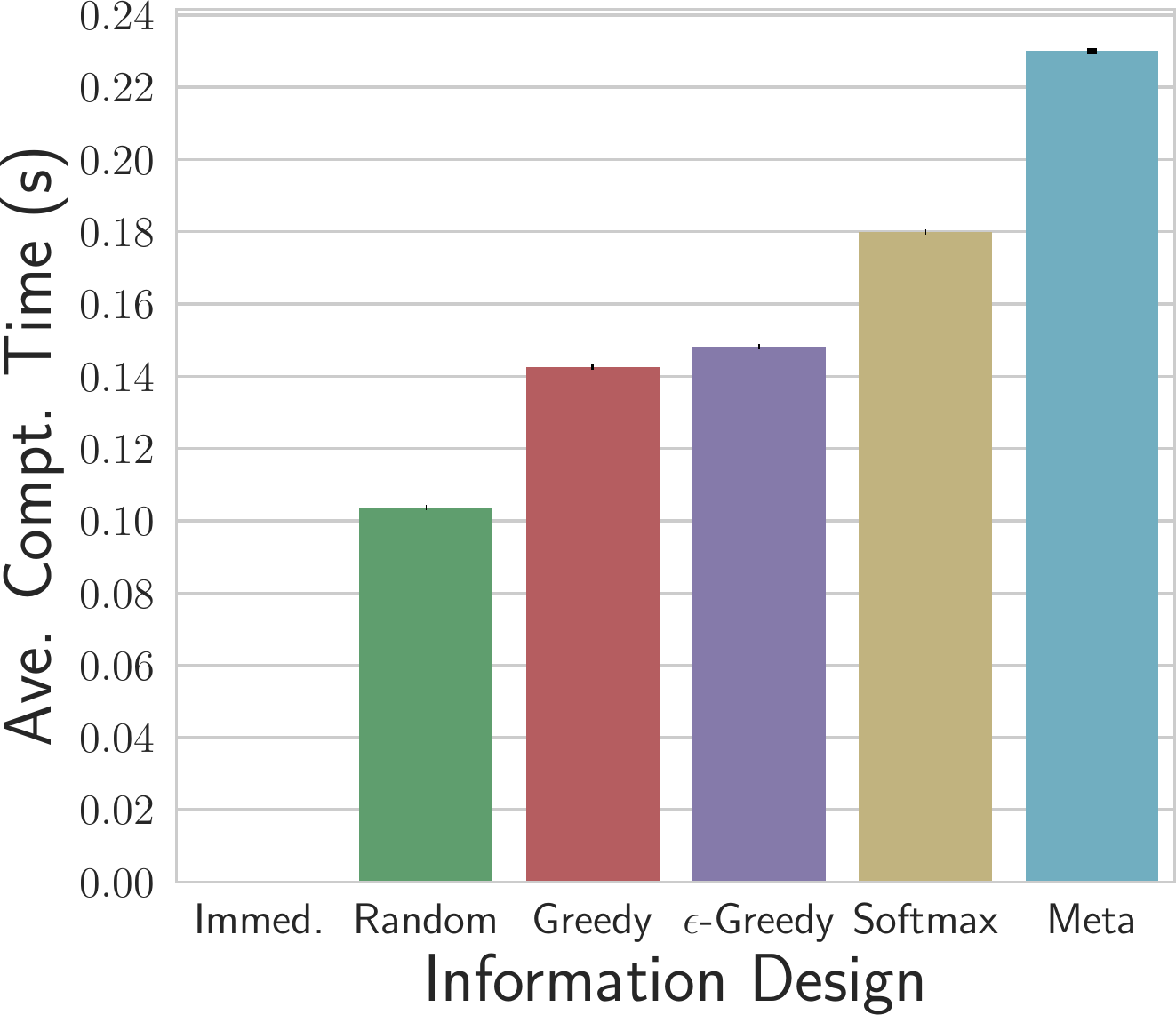}
    \caption{Computation time}
     \label{fig:time}
\end{subfigure}
\caption{A comparison of performance.}
\label{fig:comparison}
\end{figure}

Figure~\ref{fig:revenueoverall} shows the average revenue (normalized by the highest revenue achieved in all experiments) obtained in the end by each group. The actual revenue excluded the projects that failed ($M(T) < G$), while the expected revenue included all the projects regardless of whether they succeeded to meet the funding goal or not. Not surprisingly, the expected revenue of each group was significantly higher than their respective actual revenue. This is because the majority of the campaigns failed due to not having met the funding goal within the deadline (see Figure~\ref{fig:successrate}).  Among the six groups, the {\em meta} group scored the best for both the expected revenue (mean = 0.7435, std = 0.0244) and the actual revenue (mean = 0.3722, std = 0.0092), followed by the {\em softmax} group, and the {\em $\epsilon$-greedy} group.  The {\em greedy} group and the immediate group performed better than the {\em immediate} group in the expected revenue, but not in the actual revenue due to a lower success rate. The {\em random} group received the lowest scores in terms of both the expected revenue (mean = 0.3210, std = 0.0273) and the actual revenue (mean =  0.1004, std = 0.0329).

At the beginning,  the {\em immediate}, the {\em meta}, the {\em greedy}, and {\em $\epsilon$-greedy} groups  performed better than the other two (see Figure~\ref{fig:revenueovertime}). As time progressed, the {\em meta} and the {\em $\epsilon$-greedy} groups continued to lead the way until the later left behind the former at around $t=400$. The {\em $\epsilon$-greedy} group kept the second until at time $t=900$ that it was surpassed by the {\em softmax} group. One explanation is that the $\epsilon$-greedy algorithm initially encouraged exploration to a higher degree than the softmax algorithm. However, it acted more greedily than the {\em softmax} exploration over time, which was not favorable since better choices were rarely explored. The {\em meta} group used a  set of refined policies to produce more robust decisions than the others. The {\em random} group performed the worst possibly because the random algorithm completely ignored the history of backers' responses.

The {\em meta} algorithm took the most computation time (mean = 0.2300 std = 0.0290 ), while the {\em random} method required the least time (mean = 0.1037, std = 0.0034) (see Figure~\ref{fig:time}). Immediate disclosure required no additional time. 

In summary, although the {\em meta} group required the most computation time, it performed consistently the best among all the groups in terms of both actual and expected revenue. This echoes our previous findings that immediate disclosure is not always optimal in crowdfunding.

\section*{Conclusion}
%

In this paper, we present the very first study on information design where a sender interacts with multiple receivers that follow thresholding policies. Our work demonstrates that excessive information disclosure weakly shrinks the revenue in crowdfunding when backers use cutoff policies. It further shows that the widely-adopted immediate-disclosure policy is not optimal. We also present how the entrepreneur can benefit from dynamic information disclosure with appropriate heuristics. To further evaluate the performance of the heuristic algorithm, user studies and real-world deployment are needed. 

Although our analysis is in the context of crowdfunding and the DSHS  algorithm is intended to help the entrepreneur make decisions on information disclosure in crowdfunding, extensions to other domains (e.g., transportation systems,  smart grids, and online shopping) where agents typically use thresholding policies can be straightforward. For instance, online shopping marketplace can employ DSHS variants to dynamically reveal the number of products available or the number of products sold to attract potential buyers to buy the products. Further research is required to assess the variants' performance in these domains.

\begin{acks}
This work was supported by National Science Foundation through grant no. CCF-1526593. Yan was supported by National Science Foundation of China under grant no. 61602431. The authors wish to thank the anonymous reviewers for their helpful comments and suggestions.

\end{acks}
\appendix
\section{Proofs}
\label{appendix:proofs}
\textbf{Proof of Proposition 1}\\
By condition, $|s(k_{1})|  = |s(k_{2})|$. That means the revenue does not increase between  time $k_{1}$ to time $k_{2}$.  Since $k_{1} < k_{2}$, we have $T-k_{2} < T - k_{1}$, which means less time is left to achieve the fundraising goal $G$. Thus, the campaign's {\em PoS} decreases or stays the same, regardless of backers' arrivals from time $k_{1}$ to $k_{2}$. That is, $r(t, (s(k_{1}), t)) \geq r(t, (s(k_{2}), t))$. By Equation~\ref{eq:backerutility}, the utility of all the subsequently arriving backers weakly decreases, i.e, $\forall t \geq k_{2}, \forall i \in \mathcal{I}(t): u_{i}((s(k_{1}), t)) \geq  u_{i}((s(k_{2}), t))$. By Definition~\ref{def:vertical}, we have $(s(k_{1}), t) \succsim (s(k_{2}), t)$ for all $t \geq k_{2}$, and for all $i \in \mathcal{I}(t)$.\\

\noindent\textbf{Proof of Proposition 2}\\
By assumption $\forall t\in \mathcal{T}, b(t) \in \{0, 1\}$, we have that the number of arrivals from time $k_{1}$ to $k_{2}$ is:  $\sum_{k_{1}}^{k_{2}} b(t') \leq k_{2}-k_{1}$. Since $ |s(k_{2})| - |s(k_{1})| \geq (k_{2}-k_{1}) \cdot P/G$, we have that at least $(k_{2} - k_{1})$ backers have contributed from time $k_{1}$ to $k_{2}$. Without loss of generality, one can assume that each time from $k_{1}$ to $k_{2}$, at least one backer arrives at the campaign and makes a contribution. In other words, the revenue of the campaign grows faster than or equal to the arrival of backers between $k_{1}$ and $k_{2}$. Thus, the campaign's {\em PoS} increases or stays the same. That is,  $r(t, (s(k_{2}), t)) \geq r(t, (s(k_{1}), t))$. By Equation~\ref{eq:backerutility},  the utility of all the subsequently arriving backers weakly increases, i.e., $\forall t \geq k_{2}, \forall i \in \mathcal{I}(t): u_{i}(s(k_{2}), t_{2}) \geq u_{i}((s(k_{1}), t_{1}))$. By Definition~\ref{def:vertical}, we have $\forall t \geq k_{2}, \forall i \in \mathcal{I}(t): \varepsilon_{2} \succsim \varepsilon_{1}$.\\

\noindent\textbf{Proof of Lemma 1}\\
By condition, we have $\varepsilon_{2} \succsim \varepsilon_{1}$. By relation of preferences and utility\cite{chambers2016revealed},  $\varepsilon_{2} \succsim \varepsilon_{1} \iff u_{i}(t, \varepsilon_{2}) \geq u_{i}(t, \varepsilon_{1})$.  According to backers' utility function (Equation~\ref{eq:backerutility}),  $r_{i}(t, \varepsilon_{2}) \geq r_{i}(t, \varepsilon_{1})$. Depending on the order of the backer $i$'s threshold $\phi_{i}$, his belief $r_{i}(t, \varepsilon_{1})$ when given report $\varepsilon_{1}$ and the belief $r_{i}(t, \varepsilon_{2})$ given report $\varepsilon_{2}$, there are the following three cases.
\begin{enumerate}
\item[$\bullet$] $r_{i}(t, \varepsilon_{1}) \leq r_{i}(t, \varepsilon_{2}) < \phi_{i}$:  in this case, backer $i$ will not contribute to the campaign given either $\varepsilon_{2}$ or $ (\varepsilon_{1}, \varepsilon_{2})$. That is, $E(\alpha_{i} =1 | \varepsilon_{2}) = E(\alpha_{i} =1 | (\varepsilon_{1}, \varepsilon_{2})) = 0$.
\item[$\bullet$] $r_{i}(t, \varepsilon_{1}) \leq \phi_{i} \leq r_{i}(t, \varepsilon_{2})$: given report $\varepsilon_{2}$, backer $i$ will contribute to the campaign and leave the system. An additional signal about the project status cannot improve his possibility of pledging, i.e, $E(\alpha_{i} =1 | \varepsilon_{2}) \geq E(\alpha_{i} =1 | (\varepsilon_{1}, \varepsilon_{2}))$.
\item[$\bullet$] $\phi_{i} < r_{i}(t, \varepsilon_{1}) \leq  r_{i}(t, \varepsilon_{2})$: under this condition, backer $i$ will contribute to the campaign and leave the system given either of the two reports. That is, $E(\alpha_{i} =1 | \varepsilon_{2}) = E(\alpha_{i} =1 | (\varepsilon_{1}, \varepsilon_{2})) = 1$.
\end{enumerate}
Thus, $E(\alpha_{i} =1 | \varepsilon_{2}) \geq E(\alpha_{i} =1 | (\varepsilon_{1}, \varepsilon_{2}))$.\\

\noindent\textbf{Proof of Lemma 2}\\
We prove it by contradiction. Assume, to the contrary, that $\exists t\geq \max\{k_{1}, k_{2}\}$, $ j \in \mathcal{I}(t):$ $r_{j}(t, (\varepsilon_{1}, \varepsilon_{2}))$ $>  \max \{r_{j}(t, \varepsilon_{1}),  r_{j}(t, \varepsilon_{2})\}$. Depending on the order of backer $j$'s threshold $\phi_{j}$, $r_{j}(t, (\varepsilon_{1}, \varepsilon_{2}))$, and $\max \{r_{j}(t, \varepsilon_{1}),  r_{j}(t, \varepsilon_{2})\}$,  we have the following three cases:
\begin{enumerate}
\item[$\bullet$] $r_{j}(t, (\varepsilon_{1}, \varepsilon_{2})) >  \max \{r_{j}(t, \varepsilon_{1}),  r_{j}(t, \varepsilon_{2})\} \geq \phi_{j}$:  if $r_{j}(t, \varepsilon_{1}) >   r_{j}(t, \varepsilon_{2})$, we have $r_{j}(t, (\varepsilon_{1}, \varepsilon_{2})) >  r_{j}(t, \varepsilon_{1}) \geq \phi_{j}$. By utility function (Equation~\ref{eq:backerutility}), $u_{j}((\varepsilon_{1}, \varepsilon_{2})) = u_{j}(\varepsilon_{1}) = c_{j}\cdot \alpha_{j}(t)$. By the relation of preferences and utility~\cite{chambers2016revealed}, $(\varepsilon_{1}, \varepsilon_{2}) \sim \varepsilon_{1}$, where $\sim$ denotes indifferent to. This contradicts that $r_{j}(t, (\varepsilon_{1}, \varepsilon_{2})) > r_{j}(t, \varepsilon_{1})$. A similar contradiction occurs if $r_{j}(t, \varepsilon_{2}) > r_{j}(t, \varepsilon_{1})$.

\item[$\bullet$] $\phi_{j} > r_{j}(t, (\varepsilon_{1}, \varepsilon_{2})) >  \max \{r_{j}(t, \varepsilon_{1}),  r_{j}(t, \varepsilon_{2})\}$: in this case, by Equation~\ref{eq:backerutility}, $u_{j}((\varepsilon_{1}, \varepsilon_{2})) = u_{j}(\varepsilon_{1})= u_{j}(\varepsilon_{2}) = 0$.  By the relation of preferences and utility~\cite{chambers2016revealed}, $(\varepsilon_{1}, \varepsilon_{2}) \sim \varepsilon_{1} \sim \varepsilon_{2}$. This contradicts that $r_{j}(t, (\varepsilon_{1}, \varepsilon_{2})) > \max \{r_{j}(t, \varepsilon_{1}),  r_{j}(t, \varepsilon_{2})\}$. 
\item[$\bullet$] $ r_{j}(t, (\varepsilon_{1}, \varepsilon_{2})) \geq \phi_{j} >  \max \{r_{j}(t, \varepsilon_{1}),  r_{j}(t, \varepsilon_{2})\}$: by Equation~\ref{eq:backerutility}, we have $u_{j}((\varepsilon_{1}, \varepsilon_{2})) = c_{j} \cdot \alpha_{j}(t)$ and $u_{j}(\varepsilon_{1}) = u_{j}(\varepsilon_{2}) = 0$. By the relation of preferences and utility~\cite{chambers2016revealed}, $(\varepsilon_{1}, \varepsilon_{2}) \succ \varepsilon_{1}, (\varepsilon_{1}, \varepsilon_{2}) \succ \varepsilon_{2}$ and $\varepsilon_{1} \sim \varepsilon_{2}$, where $\succ$ denotes strictly preferred to.   If $\varepsilon_{1} \sim \varepsilon_{2}$, then $(\varepsilon_{1}, \varepsilon_{2}) \sim \varepsilon_{1} \sim \varepsilon_{2}$. This contradicts that $(\varepsilon_{1}, \varepsilon_{2}) \succ \varepsilon_{1}, (\varepsilon_{1}, \varepsilon_{2}) \succ \varepsilon_{2}$.
\end{enumerate}
Thus, $r_{i}(t, (\varepsilon_{1}, \varepsilon_{2})) \leq  \max \{r_{i}(t, \varepsilon_{1}),  r_{i}(t, \varepsilon_{2})\}$.\\

\noindent\textbf{Proof of Theorem 2}\\
If the information is vertical, by Lemma~\ref{lem:addivertical}, we have $E(\alpha_{i}^{'}=1)  \geq E(\alpha_{i}^{''}=1) $. If the information is horizontal, by Lemma~\ref{lem:addihorizontal}, $r_{i}(t, (\varepsilon_{1}, \varepsilon_{2})) \leq  \max \{r_{i}(t, (\varepsilon_{1})),  r_{i}(t, (\varepsilon_{2}))\}$. By Equation~\ref{eq:backerutility}, there are three cases:
\begin{enumerate}
\item[$\bullet$] $\phi_{i} \leq r_{i}(t, (\varepsilon_{1}, \varepsilon_{2})) \leq  \max \{r_{i}(t, \varepsilon_{1}),  r_{i}(t, \varepsilon_{2})\}$: in this case, backer $i$ will contribute for both conditions,  $E(\alpha_{i}^{'}=1) = E(\alpha_{i}^{''}=1) = 1 $.
\item[$\bullet$] $ r_{i}(t, (\varepsilon_{1}, \varepsilon_{2})) \leq  \max \{r_{i}(t, \varepsilon_{1}),  r_{i}(t, \varepsilon_{2})\} < \phi_{i}$: in this case, backer $i$ will not contribute for both conditions, $E(\alpha_{i}^{'}=1) = E(\alpha_{i}^{''}=1) = 0$.
\item[$\bullet$] $ r_{i}(t, (\varepsilon_{1}, \varepsilon_{2})) <  \phi_{i}  \leq \max \{r_{i}(t, \varepsilon_{1}),  r_{i}(t, \varepsilon_{2})\}$: in this case, backer $i$ will not contribute if given $(\varepsilon_{1}, \varepsilon_{2})$, i.e., $E(\alpha_{i}^{''}=1) = 0$. If given either $\varepsilon_{1}$ or $\varepsilon_{2}$, backer $i$ will either contribute or not contribute, i.e., $E(\alpha_{i}^{'}=1) \geq 0$.   
\end{enumerate}
Therefore, $E(\alpha_{i}^{'}=1) \geq E(\alpha_{i}^{''}=1)$.\\

\noindent\textbf{Proof of Lemma 3}\\
$\Longleftarrow$ If the right side holds for all $t' \leq t, i \in \mathcal{I}^{t'}$, then $\forall  t' \leq t$:  $|s(t')| - |s(t' -1)| \geq P/G = (t' - (t'-1)) \cdot P/G$. According to Proposition~\ref{ob:grow}, $\forall i \in I(t):  (s(t'), t) \succsim (s(t'-1), t)$, where $t \geq t'$. That is, immediate disclosure is always preferred by all the backers in the campaign.

$\Longrightarrow$ We prove it by contradiction. Assume, to the contrary, that  $|s(t')| - |s(t'-1)| < P/G$ such that $DP_{im}(t)$ is optimal for some $i, t'$. Without loss of generality, we let  $|s(t')| = |s(t'-1)| + |\Delta s| $, where $|\Delta s| < P/G$. Since $|s(t')| = |s(t'-1)| + \sum_{j\in \mathcal{I}(t')} \alpha_{j}(t') \cdot P/G$, where $\alpha_{i} \in \{0, 1\}$, we have $|\Delta s| = \sum_{j \in \mathcal{I}(t')} \alpha_{j}(t') \cdot P/G <  P/G$. Therefore,  $|\Delta s| = 0$, which indicates that $|s(t')| = |s(t'-1)|$. According to Proposition~\ref{ob:earliertime}, $\forall t\geq t': (s(t'-1), t) \succsim (s(t'), t)$, which contradicts the supposition that $DP_{im}$ is optimal.\\

\noindent\textbf{Proof of Lemma 4}\\
We prove it by contradiction. Assume, to the contrary, that: $DP_{im}$ is not optimal for some $d(i, t') = (s(t'), t')$. This indicates that at time $t'\geq t$,  the entrepreneur could possibly profit by delaying information disclosure. Without loss of generality, we assume that $d(i, t')_{opt} = (s(k'), t')$ is the disclosure strategy used in the optimal disclosure policy for backer $i \in \mathcal{I}(t')$ , where $k' < t'$. Depending on the relation between $t$ and $k'$, there are two cases (since $t' \geq t$) : (i) $t \leq k' < t'$;  (ii) $k' < t \leq t'$.

\begin{enumerate}
\item[$\bullet$]  $t \leq k' < t'$:  at time $t'$, the project has already succeeded, which means there is no uncertainty for backer $i$ since the campaign's PoS is 1. Therefore, delaying disclosure does not increase backer $i$'s estimate on PoS, which contradicts the proposition that $d(i,t') = (s(t'), t')$ is not optimal. 

\item[$\bullet$] $k' < t \leq t'$:  at time $t'$, backer $i$'s estimate $r_{i}(t', d(i, t')_{opt}) \leq 1$, while $r_{i}(t', d(i, t')) = 1$. That is, $d(i, t') \succsim d(i, t')_{opt} $ ($\forall i \in \mathcal{I}(t')$), which contradicts the proposition that $DP_{im}$ is not optimal.
\end{enumerate}

\section{Expert Algorithms}
\label{apendix:algo}
\noindent\textbf{Random Selection}\\
\label{app:random}
The random algorithm picks the project status $s(k_{sel})$ at random from $s(k) \in H_{i}(t)$ with equal probability.  The expected increase of revenue $\Upsilon(s(k), i, t)$ is computed by averaging the revenue received when the entrepreneur disclosed the project status $s(k)$.\\

\noindent\textbf{Greedy Selection}\\
\label{app:greedy}
The greedy selection algorithm chooses the project status $s(k_{sel})$ based on the empirical responses that the entrepreneur has received from the backers, using a one-step-look-ahead approach.

At time $t$, given the disclosure strategy $d(i, t) = (s(k), t)$, the entrepreneur establishes a historical belief $\Upsilon_{old}(s(k), i, t)$ of the expected increase in the revenue, where $i \in \mathcal{I}(t)$, and $s(k) \in H_{i}(t)$:
\begin{equation}
\label{eq:hisbelief}
\Upsilon_{old}(s(k), i, t) = \frac{1}{n_{k}(t)} \cdot \sum_{t' = 1}^{t-1} \sum_{j \in \mathcal{I}(t')} \frac{\alpha_{j}(t')}{\eta(t')/\eta(t)} \;,
\end{equation}
where $n_{k}(t)$ denotes the times that $s(k)$ has been revealed to backers up to time $t$.  $\alpha_{j}(t')$ is backer $j$'s action given disclosure strategy $(s(k), t')$.  $\eta(t') = |s(t')|/t'$, and $\eta(t) = |s(t)|/t$ are the revenue growth rates up to time $t'$ and $t$, respectively. Note that $\Upsilon_{old} = 0$ if $n_{k}(t) = 0$ or $t=1$. 

The historical belief represents the entrepreneur's estimate of the average revenue she receives by revealing project status $s(k)$ to the backers, with discounting of the revenue growth rates. It is a rough estimate of revenue increase that $d(i, j) = (s(k), t)$ brings.

At time $t$, the entrepreneur's temporal belief $\Upsilon_{tmp}$ of the expected increase in the revenue given the disclosure decision $d(i, t) = (s(k), t)$,  is determined as follows:
\begin{equation}
\label{eq:tempbelief}
\Upsilon_{tmp}(s(k), i, t) = \sum_{j \in \mathcal{I}(t-1)}\frac{\alpha_{j}(t-1)}{|\mathcal{I}(t-1)|} \;,
\end{equation}
where $\mathcal{I}(t-1)$ denotes the set of backers in the campaign at time $t-1$ and $\alpha_{j}(t-1)$ is backer $j$'s action at time $t-1$. If $|\mathcal{I}(t-1)| = 0$ or $t=1$, then $\Upsilon_{tmp} = 0$. The temporal belief captures the latest decisions of the backers that will most probably stay in the campaign at time $t$.

The entrepreneur estimates the belief of the expected increase of revenue, given $d(i, t) = (s(k), t)$ for backer $i \in \mathcal{I}(t)$,  with the following equation:
\begin{equation}
\label{eq:beliefrev}
\Upsilon(s(k), i, t) = (1-\lambda)  \Upsilon_{old}(s(k), i, t) +\lambda \Upsilon_{tmp}(s(k), i, t) \;,
\end{equation}
where $\lambda \in [0, 1] $ is the learning rate (we use $\lambda = 0.1$).

The greedy algorithm then selects the project status $s(k_{sel})$ by using the following equation:
\begin{equation}
\label{eq:greedy}
s(k_{sel}) = \argmax_{s(k) \in H} \Upsilon(s(k), i, t) \; .
\end{equation}
The probability for selects the project status $s(k_{sel})$ is 1 and 0 for others.\\

\noindent\textbf{$\epsilon$-Greedy Exploration}\\
\label{app:egreedy}
In $\epsilon$-greedy exploration,  with probability $\epsilon$ the algorithm  selects a random choice $s(k)$. Otherwise, with probability $1-\epsilon$ it  selects the greedy choice determined in Equation~\ref{eq:greedy}.
\begin{equation}
Pr(s(k)) =
  \begin{cases}
   1-\epsilon+ \frac{\epsilon}{n_{k}(t)} \; ,      & \quad \text{if  Equation~\ref{eq:greedy}}\\
    \frac{\epsilon}{n_{k}(t)} \; ,  & \quad \text{otherwise}\\
  \end{cases}
\end{equation}
where $n_{k}(t)$ is the times that $s(k)$ has been reveled to backers up to time $t$,  $\epsilon = c/n_{k}(t)$, and $c \in [0, 1]$ is a constant. Note that $\epsilon = c$ if $n_{k}(t) = 0$. \\

\noindent\textbf{Softmax Exploration}\\
\label{app:softmax}
Softmax selects the choice using a Boltzmann distribution~\cite{ross1996stochastic}. At time $t$, the algorithm selects choice $s(k)$ with the probability:
$Prob(s(k)) = \frac{e^{\Upsilon(s(k),i, t)/\tau}}{\sum_{s(k) \in H_{i}(t)} e^{\Upsilon(s(k),i, t)/ \tau}}$,
where  $\tau = \max \{\mu, C^{t}/ \log n_{k}(t) \}$ is the temperature parameter.
Here,  $\mu = 0.0001$,  $\tau = 1$ when $n_{k}(t)=0$, and $C^{t}$ is determined by: 
$C^{t}  =$ $ \max_{s(k_{1}), s(k_{2}) \in H_{i}(t)} |\Upsilon(s(k_{1}), i, t) - \Upsilon(s(k_{2}), i, t)|$.
The softmax exploration selects each choice with a probability that is proportional to the average $\Upsilon$. 
\bibliographystyle{ACM-Reference-Format}  
\bibliography{crowdfunding}  

\end{document}